# Efficient Measuring of Readability to Improve Documents Accessibility for Arabic Language Learners

Sadik Bessou, Ghozlane Chenni
Department of Computer Science, Faculty of Sciences
University of Ferhat Abbas Sétif 1
Algeria
{bessou.s@univ-setif.dz} {cghozlene@gmail.com}

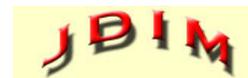

**ABSTRACT:** This paper presents an approach based on supervised machine learning methods to build a classifier that can identify text complexity in order to present Arabic language learners with texts suitable to their levels.

The approach is based on machine learning classification methods to discriminate between the different levels of difficulty in reading and understanding a text. Several models were trained on a large corpus mined from online Arabic websites and manually annotated.

The model uses both Count and TF-IDF representations and applies five machine learning algorithms; Multinomial Naïve Bayes, Bernoulli Naïve Bayes, Logistic Regression, Support Vector Machine and Random Forest, using unigrams and bigrams features. With the goal of extracting the text complexity, the problem is usually addressed by formulating the 'level identification' as a classification task.

Experimental results showed that n-gram features could be indicative of the reading level of a text and could substantially improve performance, and showed that SVM and Multinomial Naïve Bayes are the most accurate in predicting the complexity level. Best results were achieved using TF-IDF Vectors trained by a combination of word-based unigrams and bigrams with an overall accuracy of 87.14% over four classes of complexity.

**Subject Categories and Descriptors: I.2.7 [Natural Language Processing]:** *Discourse, Language parsing and understanding, Text analysis*. **I.7 [Document and Text Processing]:** Document Preparation, Document and Text Editing.

## 1. Introduction

Reading is a crucial ability for humans. It is indispensable to learn and master knowledge, and to obtain information from the world, but not all content is readable by all.

While reading a document, we might struggle with unfamiliar words, which make a text difficult to understand. Text readability estimates the degree of difficulty of a text, and measures its appropriateness to particular readers. It is widely used in the education field to select texts that best match a learner's reading level and to support educators in drafting textbooks and curricula to suit each age of students. Ascertaining the readability of curricula is an important step toward optimizing the effectiveness of the educational progress.

Measuring the readability of the text is to assign a rating indicating how difficult an existing text is. It indicates the required literacy levels needed to understand the information presented in a text. Therefore, the complexity level would be used to select the text to read.

Text readability enables selecting a document with an appropriate difficulty among many reading documents. A suitable text could help readers to more understand the information conveyed within a text. Furthermore, text readability assists content creators in producing or adapting texts suitable for the target audience [1]. Therefore, predicting readability is important for writers, educators and learners.

Moreover, text readability is a fundamental issue on e-learning, online question-answering, Web hosting, searching and browsing [2].

One important application of the complexity analysis is the text simplification[3], [4]. It involves decreasing the complexity of vocabulary by selecting words or phrases in a text that are considered complex for the reader. Therefore, texts are more accessible to individuals with low literacy levels, people with reading difficulties including dyslexics and people with comprehension disabilities.

In this paper, we address text readability by focusing on the problem of predicting the difficulty of text using machine learning techniques. There are several studies about English and European languages, but not so many experiments on Arabic. Thus, we propose to use machine-learning techniques to learn a supervised classifier to determine Arabic text difficulty.

## 2. Related Work

In this section we provide a review of the major works that have been devoted to text readability. Researchers from different disciplines proposed different approaches to classify or score texts according to their complexity.

There are numerous studies on computing text complexity, most of them focusing on English and similar initiatives are often missing in other languages like Arabic.

[5] proposed a set of features for sentence extraction using summarization techniques. The experiments showed that learner dependent features improved the readability of the text through summary as a text preview.

[6] evaluated the impact of removing stop words on the performance of readability assessment of Thai text. The Authors used mutual information method to select top-ranking terms and a TF-IDF value vector of the selected terms was computed for each document. SVM was used to generate prediction models for assessing the readability using these feature vectors. Experimental results showed that the F-measure could reach 0.87.

[7] investigated the role of text length in assessing the Vietnamese text readability. The experiment results showed that the features related to the text length have a huge impact on readability assessment for textbooks.

In another study, [8] presented a method for assessing the readability of literary texts in the Vietnamese textbooks based on specific features of Vietnamese language. The authors used SVM to classify texts by readability. The experimental results showed that the proposed method remarkably improves the accuracy of the assessing and the features used were valuable.

[9] investigated three types of lexical chains: exact, synonymous, and semantic to estimate their overall capacity to distinguish between easy and difficult text. They tested the usefulness of features at sentence-level of simplification. The Authors found that lexical chain features performed significantly better than the bag-of-words baseline across a range of classifiers: Logistic Regression, Naïve Bayes, Decision Trees, Linear and RBF kernel SVM, and Random Forest with the best classifier achieving an accuracy of 90%.

[10] developed an automated readability assessment estimator based on supervised learning algorithms over German texts at the sentence level. The authors extracted 73 linguistic features and employed feature engineering to select informative features. The results obtained depicted that Random Forest Regressor yielding better results.

[11] proposed a neural network-based approach to predict a readability score for a given sentence in the Kannada language without the use of any predefined word list. The Author used a regression model to predict a score that corresponds to the class labels. The model consists of 4 hidden layers, 1 input, and 1 output layer. Correlation achieved 0.89 with 30 sentences in a paragraph.

[12] compared the text readability and syntactic complexity of English news texts among eight ASEAN countries, England and America. The Authors used the hierarchical clustering method to classify the English texts of different countries into six different levels according to the difficulty of the text.

Arabic readability research suffers from limited and a little attention from the Arab research community. One of the earlier studies on Arabic readability was the work of [13] where authors proposed a system for automating the readability measurement of Arabic text. The system assigns a given Arabic text a readability level (easy, medium, difficult). No details were given about the approach or the evaluation.

[14] evaluated the informativeness of lexical, morphological, and semantic features in determining the readability of texts geared towards learners of Arabic as a foreign language. The Authors gathered low-complexity features and tested common classification algorithms.

Results indicated that a small set of easily computed fea-tures such as morpheme counts, type and token counts, part-of-speech, and various measures of sentence and document length can be indicative of the reading level of a text.

From this researches we note that many features have been evaluated to measure text difficulty, ranging from features that focus on individual words or sentences to entire documents. But, the absence of morphological features in readability measurements for languages other than English is the main critical observation when it comes to analyzing the previous works.

Several studies confirmed that the length of the word, classically considered as an important cue for complexity, is not a good feature for the classifiers. On the other hand, frequency of the word in reference corpora is an informative feature, especially when combining frequency from simple and general corpora [15].

In a recent study, [16] found that the higher the difficulty level of the text, the richer the words in the text and the more diverse the expressions. This motivates the importance of using n-gram features to distinguish between easy and difficult texts. These features are crucial when assessing the readability of languages rich in morphology, such as Arabic. In this research, we examine the impact of n-gram features on computing readability from a supervised point of view.

### 3. Data and Methods

#### 3.1. Dataset

Our dataset consists of Arabic texts that are manually collected from the web, on different topics. The dataset covers four major complexity levels. We collected 39,792 documents where the texts are annotated with readability level. We made four distinct groups for each difficulty level: Easy, Medium, Difficult, and Very difficult. The distribution of the texts over the levels is shown in table 1 and figure 1.

| Category | #Documents | #Tokens |
|---|---|---|
| Easy | 6,691 | 103,032 |
| Medium | 8,844 | 125,484 |
| Difficult | 9,326 | 125,550 |
| Very difficult | 14,931 | 121,388 |
| **Total** | **39,792** | **475,454** |

Table 1. Number of documents and tokens in the corpus

The four categories were collected and annotated with the following criteria:

**Easy Documents:** Include documents written for chil-

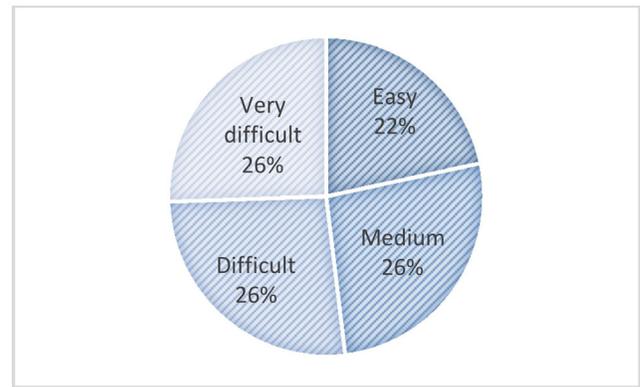

Figure 1. The distribution of the texts over the four levels of difficulty

dren or for people having primary education. These documents were mainly collected from primary school textbooks, primary sample essays, fairytales, storybooks for children, etc.

**Medium Documents:** These documents are written for middle and high school students, or people with high school education. Documents in this category were collected from textbooks, student magazine articles, and news articles.

**Difficult Documents:** Include documents written for college students, specialized documents, scientific papers, etc. These documents were collected from university textbooks, specialized documents, political theory articles, language and literary articles, law and religious documents, etc.

**Very Difficult Documents:** These documents are understandable by linguists and historians where the language is different than modern standard Arabic (MSA). These documents were collected from the pre-Islamic era.

#### 3.2. Preprocessing

The aim of this phase is to improve the text classification by removing worthless information. The scrapped data necessitate pre-processing since noisy and worthless information data can decrease the efficiency of the system. To improve the quality of the input data and to ensure the quality of the features being extracted, we cleaned up the unwanted content by performing the following processes as mentioned in figure 2:

**Tokenizing**
It consists of splitting the texts into words, which are the usual units of processing for morphological analyzers. In this step, we normalized our data based on white space, excluding all non-Unicode characters.

**Filtering**
The purpose of filtering is to remove the character sequences that may be noisy and consequently affect the quality of data. After converting text corpus into UTF"8 encodings, we have cleaned up the texts by stripping punctuation marks, special characters, non-Arabic characters, dates, time, numbers, single letters, links, diacritics, etc.

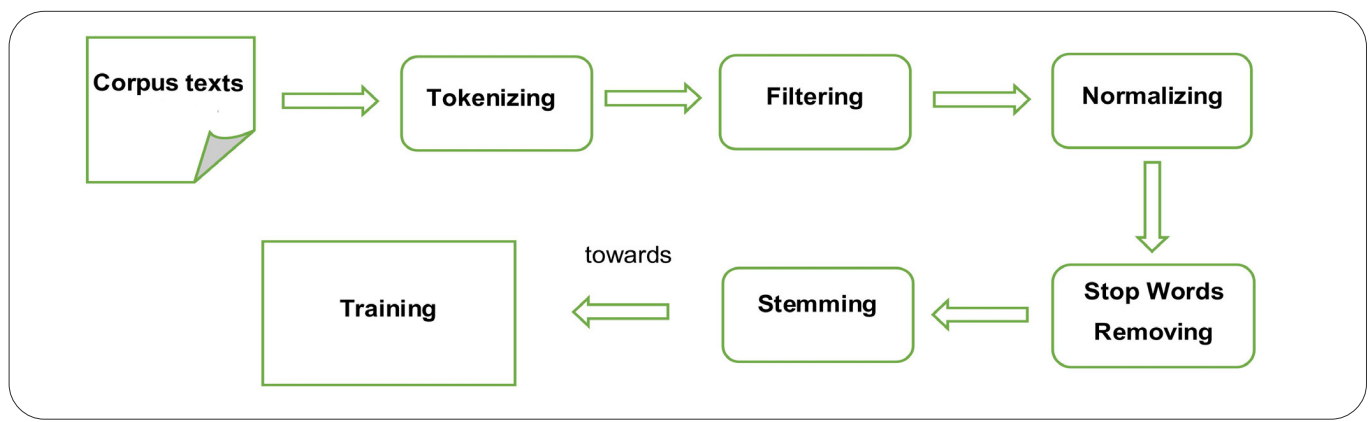

Figure 2. Preprocessing steps before training

**Normalizing**
In this step, we replaced specific letters within the word with other letters according to a predefined set of rules; i.e., the unification of characters. Some writing forms (Hamza and Alif) need normalization, we have for instance converted "أ", "إ", "آ" into "ا" since most of the Arabic texts neglect the addition of Hamza on Alif.

**Stop Words Removing**
Stop words (pronouns, conjunctions, prepositions, and names) have little effect on identifying classes of texts, and therefore may be removed in advance. They are considered as valueless for taking them as features. Therefore, we have removed all stop words.

**Stemming**
It consists of removing affixes from words, and reducing these words to their roots. It can significantly improve the efficiency of the classification by reducing the number of terms being input to the classification.

Many stemming methods have been developed for the Arabic language. The two most widely used stemming methods are:

**1. The heavy stemming:** Allows transforming each surface Arabic word in the document into its root [17].

**2. The light stemming:** Allows removing prefixes and suffixes [18].

In this work, we used light stemming. That does not reduce a word to its proper root, but it removes only prefixes and suffixes from words [19], as the removal of infixes can change the word meaning completely and consequently the complexity level.

**3.3. Training and Test Datasets**
In this step, we labeled documentcorpus, whose total number is 39,792, and then we split the data into a training set and test set. The training was performed on 80% of the data, which consists of 31,833 documents, and 20% were kept for testing which consists of 7,959 documents. The system takes the pre-classified documents and applies the selected machine-learning algorithm to train the system and build a classification model, which will be used to classify new documents by readability level.

**3.4. Proposed Model**
Our proposed system is classified as a data driven approach that automates the task of assigning a given Arabic text to a readability level as a class (Easy, Medium, Difficult, and Very difficult). Each level has distinct words used only in that level, but not in other levels. Therefore, the approach focuses on word-based n-grams using various classification algorithms for assessing the lexical complexity. To do so, we propose to use a set of word n-gram features and perform a classification task using supervised machine learning methods. We study the performance of different algorithms in this classification task and determine which of the defined features are more important to determine the lexical complexity.

We extract two lengths of n-grams, 1-2 grams. These n-grams are used as features in the bag of words model, which builds a term-document matrix by assigning a weight to every term that appears in each document. Many schemes of this model can be used. In our case, we use Count Vectors based on 1-2 grams (binary weights) and Term-FrequencyInverse-Document-Frequency (TF-IDF) Vectors based on 1-2 grams (sophisticated weights). For each complexity level, we train a word-level language model.

We formulate the task as a multi-class classification problem, where each complexity level is a separate class. Given a collection of documents and associated levels, we consider a supervised system to predict the levels of the documents, $f: D_i \rightarrow L_i$. It assigns to each document $D_i$, the level $L_i$ that maximizes its conditional probability score $argmax_i P(L_i \backslash D_i)$. To understand how the features perform across multiple different learning approaches, and to identify which classifiers work best for this problem domain, we investigated five machine learning algorithms: Multinomial Naïve Bayes, Bernoulli Naïve Bayes, Logistic Regression, Support Vector Machine and Random Forest trained on labeled data to document level. These classifiers are suitable for classification tasks with discrete features; word or character counts representation for text classification [20]. Four classes are predicted (Easy, Medium, Difficult, and Very difficult). The goal of the experiments is to find the highest accuracy using different classifiers.

We used default settings for Logistic Regression and Multinomial Naïve Bayes. For the Support Vector Machine we changed the number of iterations to 1500.

In the Logistic Regression and the Support Vector Machine, we used L1 and L2 regularization, which can be added to the algorithm to ensure that the models do not overfit its data. The L1 regularization norm is the sum of the absolute differences between the estimated and target values, while the L2 regularization norm is the sum of the square of the differences between estimated and target values. The regularization value of 1.0 has been used for class weighting. For Multinomial Naïve Bayes, we used Laplace smoothing regularization method.

## 4. Results and Discussion

We have carried out two different experiments using n-grams with $n$ in $\{1, 2\}$ on either Count or TF-IDF Vectors. The results of the different classifiers using all set of features are compared to find out which classifier is most accurate in the task.

### 4.1. Results using Count Vectors

Tables 2-4 show the results using Count Vectors. Accuracy, Precision, Recall, and F-measure are the measures of performance used in the experiments. The tables present the performance of the used classification algorithms. We notice that Multinomial Naïve Bayes achieved highest scores over all experiments with the best accuracy of 86.47%. All accuracies of Multinomial Naïve Bayes are above 76%.

#### 4.1.1. Using Unigrams

Table 2 shows the accuracy, precision, recall and F1-score of the five algorithms with the use of unigram feature.

As it shown in table 2, Multinomial Naïve Bayes achieved the best accuracy with 85.20% followed by Logistic Regression with 84.74%.

#### 4.1.2. Using Bigrams

The results using bigrams feature are shown in table 3. We notice that multinomial Naïve Bayes obtained the best performance with an accuracy of 76.74%.

#### 4.1.3. Using Unigrams and Bigrams

The results using combination of unigrams and bigrams are shown in table 4.

The best classifier in the third experiment is similarly Multinomial Naïve Bayes with 86.47%.

From the three experiments, we notice that the combination of unigrams and bigrams provided the best accuracies. Multinomial Naïve Bayes outperformed other algorithms, achieving the highest accuracy rate of 86.47%, followed by Logistic Regression with 84.81%, then Support Vector Machine with 83.89%. The accuracy decreased to 72.97% for Random forest and the worst classifier was Bernoulli Naïve Bayes which has an accuracy of 56.46 %.

### 4.2. Results using TF-IDF Vectors

Tables 5-7 show the results using TF-IDF Vectors. The tables present the performance of the used classification algorithms. We notice that SVM achieved the highest scores overall experiments with the best accuracy of 87.14%. All accuracies of SVM are above 74%.

#### 4.2.1. Using Unigrams

Table 5 shows the accuracy, precision, recall and F1-score of the five algorithms with the use of the unigram feature.

As it is shown in table 5, SVM achieved the best accuracy with 87.03%, followed by multinomial Naïve Bayes with 85.87%.

#### 4.2.2. Using Bigrams

The results using bigrams feature are shown in the table 6. We notice that SVM obtained the best performance with an accuracy of 74.41%.

#### 4.2.3. Using Unigrams and Bigrams

The results using a combination of the unigrams the and bigrams are shown in table 7.

The best classifier in the third experiment was similarly SVM with 87.14%.

From the three experiments, we notice that the combination of unigrams and bigrams provided the best accuracies. SVM outperformed other algorithms, achieving the highest accuracy rate of 87.14%, followed by Multinomial Naïve Bayes with 84.78%, then Logistic Regression with 83.72%. The accuracy decreased to 72.58% for Random forest and the worst classifier was Bernoulli Naïve Bayes which has an accuracy of 55.63 %. Over the six experiments we notice that SVM performs better with TF-IDF Vectors than with Count Vectors. Multinomial Naïve Bayes performs better with Count Vectors than with TF-IDF Vectors. For all classifiers, the accuracy is better using a combination of unigrams and bigrams features than using each one alone.

| Algorithm | Accuracy | Precision | Recall | F1-score |
|---|---|---|---|---|
| **Multinomial Naïve Bayes** | **85.20%** | 0.84 | **0.85** | **0.85** |
| **Bernoulli Naïve Bayes** | 74.67% | 0.86 | 0.69 | 0.73 |
| **Logistic Regression** | 84.74% | **0.87** | 0.82 | 0.84 |
| **Support Vector Machine** | 83.56% | 0.84 | 0.82 | 0.83 |
| **Random Forest** | 73.40% | 0.73 | 0.73 | 0.73 |

Table 2. Performance measures with Count Vectors using unigrams

| Algorithm | Accuracy | Precision | Recall | F1-score |
|---|---|---|---|---|
| **Multinomial Naïve Bayes** | **76.74%** | 0.83 | **0.72** | **0.75** |
| **Bernoulli Naïve Bayes** | 44.74% | 0.85 | 0.33 | 0.27 |
| **Logistic Regression** | 65.80% | **0.84** | 0.58 | 0.62 |
| **Support Vector Machine** | 69.35% | 0.83 | 0.63 | 0.67 |
| **Random Forest** | 55.24% | 0.75 | 0.59 | 0.58 |

Table 3. Performance measures with Count Vectors using bigrams

| Algorithm | Accuracy | Precision | Recall | F1-score |
|---|---|---|---|---|
| **Multinomial Naïve Bayes** | **86.47%** | 0.86 | **0.86** | **0.86** |
| **Bernoulli Naïve Bayes** | 56.46% | 0.85 | 0.47 | 0.48 |
| **Logistic Regression** | 84.81% | **0.88** | 0.82 | 0.84 |
| **Support Vector Machine** | 83.89% | 0.85 | 0.82 | 0.83 |
| **Random Forest** | 72.97% | 0.74 | 0.72 | 0.72 |

Table 4. Performance measures with Count Vectors using unigrams and bigrams

| Algorithm | Accuracy | Precision | Recall | F1-score |
|---|---|---|---|---|
| **Multinomial Naïve Bayes** | **85.87%** | 0.87 | **0.84** | **0.85** |
| **Bernoulli Naïve Bayes** | 74.59% | 0.86 | 0.69 | 0.73 |
| **Logistic Regression** | 83.69% | **0.86** | 0.81 | 0.83 |
| **Support Vector Machine** | 87.03% | 0.88 | 0.86 | 0.87 |
| **Random Forest** | 74.87% | 0.75 | 0.74 | 0.74 |

Table 5. Performance measures with TF-IDF Vectors using unigrams

The classification accuracy increases considerably in detecting text difficulty this could be explained by the importance of *n*-gram features and their impact on improving the accuracy.

| Algorithm | Accuracy | Precision | Recall | F1-score |
|---|---|---|---|---|
| **Multinomial Naïve Bayes** | **67.91%** | 0.84 | **0.61** | **0.64** |
| **Bernoulli Naïve Bayes** | 44.26% | 0.85 | 0.32 | 0.26 |
| **Logistic Regression** | 62.91% | **0.83** | 0.55 | 0.58 |
| **Support Vector Machine** | 74.41% | 0.83 | 0.69 | 0.73 |
| **Random Forest** | 55.25% | 0.79 | 0.63 | 0.66 |

Table 6. Performance measures with TF-IDF Vectors using bigrams

| Algorithm | Accuracy | Precision | Recall | F1-score |
|---|---|---|---|---|
| **Multinomial Naïve Bayes** | **84.78%** | 0.88 | **0.82** | **0.84** |
| **Bernoulli Naïve Bayes** | 55.63% | 0.84 | 0.46 | 0.47 |
| **Logistic Regression** | 83.72% | **0.86** | 0.82 | 0.83 |
| **Support Vector Machine** | 87.14% | 0.87 | 0.86 | 0.87 |
| **Random Forest** | 72.58% | 0.73 | 0.72 | 0.72 |

Table 7. Performance measures with TF-IDF Vectors using unigrams and bigrams

Based on the experimental results, it can be concluded that Multinomial Naïve Bayes performs substantially with Count Vectors when SVM delivers good performance with TF-IDF Vectors. This suggests that the fine choice of data representation is crucial and produces better results. Upon examining the full scores, we notice that both experiments yield strong comparable results that range between 74.41% and 87.14% for different features. Furthermore, most scores with TF-IDF Vectors representation tend to be better than scores with Count Vector representation.

## 5. Conclusion

In this paper, we used machine learning to detect text complexity levels to improve the document's accessibility for Arabic learners. Measuring the readability of the text is one of the essential concerns for helping learners. Several models were trained for complexity identification. We used various classifiers and features to find the best models to predict the level. The results showed that combination of n-gram features could substantially improve performance. Additionally, we noticed that the kind of data representation could provide a significant performance boost compared to simple representation.

The different algorithms achieved good scores, with SVM is the best. The overall performance is promising. However, some improvements are feasible by using a larger corpus covering additional domains, exploring deeper features like syntactic and semantic features and deal ing with more fine-grained difficulty levels.

## References

[1] Zambrano, J. O., Tapia, E. V. (2018). Reading comprehension in university texts: the metrics of lexical complexity in corpus analysis in Spanish. *In*: *International Conference on Computer and Communication Engineering* (p. 111-123). Springer, Cham, (October).

[2] Zhang, Lixiao, Liu, Zaiying., Ni, Jun. (2013). Feature-based assessment of text readability. *In*: 2013 *Seventh International Conference on Internet Computing for Engineering and Science*, p. 51-54. IEEE.

[3] Silpa, K. S., Irshad, M. (2018). Lexical Simplification of Complex Scientific Terms. *In*: 2018 International Conference on Emerging Trends and Innovations In Engineering and Technological Research (ICETIETR), p. 1-5. IEEE.

[4] Wibowo, Satrio, Muhammad., Romadhony, Ade., Siti Sa'adah. (2019). Lexical and Syntactic Simplification for Indonesian Text. *In*: 2019 International Seminar on Research of Information Technology and Intelligent Systems (ISRITI), p. 64-68. IEEE.

[5] Nandhini, Kumaresh., Ramakrishnan Balasundaram, Sadhu. (2012). Significance of learner dependent features for improving text readability using extractive summarization. *In*: *2012 4th International Conference on Intelligent Human Computer Interaction (IHCI)*, p. 1-5. IEEE.

[6] Daowadung, Patcharanut., Yaw-Huei Chen. (2012). Stop word in readability assessment of Thai text. *In*: *2012 IEEE 12th International Conference on Advanced Learning Technologies*, p. 497-499. IEEE.

[7] Luong, An-Vinh, Nguyen, Diep., DienDinh. (2017). Examining the text-length factor in evaluating the readability of literary texts in Vietnamese textbooks. *In*: *2017 9th International Conference on Knowledge and Systems Engineering (KSE)*, p. 36-41. IEEE.

[8] Luong, An-Vinh, Nguyen, Diep., DienDinh. (2018). Assessing the Readability of Literary Texts in Vietnamese Textbooks. *In*: *2018 5th NAFOSTED Conference on Information and Computer Science (NICS)*, p. 231-236. IEEE.

[9] Mukherjee, Partha, Leroy, Gondy., Kauchak, David. (2018). Using Lexical Chains to Identify Text Difficulty: A Corpus Statistics and Classification Study. *IEEE journal of biomedical and health informatics,* 23 (5) 2164-2173.

[10] Naderi, Babak, SalarMohtaj, Karan, Karan., Möller, Sebastian. (2019). Automated Text Readability Assessment for German Language: A Quality of Experience Approach. *In*: *2019 Eleventh International Conference on Quality of Multimedia Experience (QoMEX)*, p. 1-3. IEEE.

[11] Narasinh, Vishwaas. (2019). Readability Analysis of Kannada Language. *In*: *2019 1st International Conference on Advances in Information Technology (ICAIT)*, p. 45-49. IEEE.

[12] Zhang, Yusha, Lin, Nankai., Jiang, Shengyi. (2019). A Study on Syntactic Complexity and Text Readability of ASEAN English News. *In*: 2019 *International Conference on Asian Language Processing (IALP)*, p. 313-318. IEEE.

[13] Al-Ajlan, Amani, A., Hend, S., Al-Khalifa, Abdul Malik S., Al-Salman. (2008). Towards the development of an automatic readability measurements for Arabic language. *In*: 2008 *Third International Conference on Digital Information Management*, p. 506-511. IEEE.

[14] Saddiki, Hind, Bouzoubaa, Karim., Cavalli-Sforza, Violetta. (2015). Text readability for Arabic as a foreign language. *In*: 2015 *IEEE/ACS 12th International Conference of Computer Systems and Applications (AICCSA)*, p. 1-8. IEEE.

[15] Wilkens, R., DallaVecchia, A., Boito, M. Z., Padró, M., Villavicencio, A. (2014). Size does not matter. Frequency does. A study of features for measuring lexical complexity. *In*: Ibero-American Conference on Artificial Intelligence (p. 129-140). Springer, Cham, (November).

[16] Wang, Huiping., Yang, Lijiao., Xiao, Huimin. (2019). Construction of Quantitative Index System of Vocabulary Difficulty in Chinese Grade Reading. *In*: 2019 *International Conference on Asian Language Processing (IALP)*, p. 480-486. IEEE.

[17] Khoja, Shereen., Garside, Roger. (1999). Stemming arabic text. Lancaster, UK, Computing Department, Lancaster University.

[18] Larkey, Leah, S., Margaret, E., Connell. (2001). Arabic information retrieval at UMass in TREC-10. In TREC. 2001.

[19] Bessou, Sadik., Touahria, Mohamed. (2014). An accuracy-enhanced stemming algorithm for Arabic information retrieval. *Neural Network World*, 24 (2) 117.

[20] Schütze, Hinrich., Christopher, D., Manning, Raghavan, Prabhakar. (2008). Introduction to information retrieval. Vol. 39. Cambridge: Cambridge University Press.